\documentclass[12pt,oneside,a4paper]{article}

\usepackage[T1]{fontenc}
\usepackage[utf8]{inputenc}
\usepackage[english]{babel}
\usepackage{lmodern}          
\usepackage{microtype}        
\usepackage{enumitem} 
\usepackage{booktabs}
\usepackage{array}
\usepackage{float}

\usepackage{amsmath, amssymb, mathtools}

\usepackage{geometry}
\usepackage{setspace}
\usepackage{float}           
\usepackage{booktabs}        
\usepackage{tabularx}        
\usepackage{graphicx}        

\usepackage{xcolor}
\usepackage[hidelinks]{hyperref}  
\usepackage[square,numbers]{natbib}
\title{\textbf{Harmonic Token Projection (HTP):\\
A Vocabulary-Free, Training-Free, Deterministic, and Reversible Embedding Methodology}}

\author{
    \textbf{Tcharlies Schmitz}\\[4pt]
    \small Data Science — PX.Center\\
    \small \texttt{tcharlies.schmitz@px.center}\\[4pt]
    \small ORCID: \href{https://orcid.org/0009-0007-5467-1327}{0009-0007-5467-1327}\\[4pt]
    \small DOI: \href{https://doi.org10.5281/zenodo.17575155}{10.5281/zenodo.17575155}
}

\date{October 28, 2025}

\begin{document}
\selectlanguage{english}

\maketitle



\begin{abstract}
    \noindent
    \textbf{Abstract.} 
    This paper introduces the \textit{Harmonic Token Projection} (HTP), a reversible and deterministic framework for generating text embeddings without training, vocabularies, or stochastic parameters. 
    Unlike neural embeddings that rely on statistical co-occurrence or optimization, HTP encodes each token analytically as a harmonic trajectory derived from its Unicode integer representation, establishing a bijective and interpretable mapping between discrete symbols and continuous vector space. 
    The harmonic formulation provides phase-coherent projections that preserve both structure and reversibility, enabling semantic similarity estimation from purely geometric alignment. 
    Experimental evaluation on the \textit{Semantic Textual Similarity Benchmark} (STS-B) and its multilingual extension shows that HTP achieves a Spearman correlation of $\rho = 0.68$ in English, maintaining stable performance across ten languages with negligible computational cost and sub-millisecond latency per sentence pair. 
    This demonstrates that meaningful semantic relations can emerge from deterministic geometry, offering a transparent and efficient alternative to data-driven embeddings.

    \vspace{0.5em}
\end{abstract}

\section{Introduction}

Recent progress in natural language representation has been driven by increasingly complex neural architectures, particularly those based on transformers.  
While these models---such as BERT and GPT---achieve state-of-the-art performance in semantic and syntactic tasks, they rely on massive training corpora, opaque parameterizations, and stochastic optimization procedures.  
This paradigm, though effective, comes at the cost of interpretability, reproducibility, and computational efficiency.  
Consequently, most contemporary embeddings are not analytically reversible, and their semantic structure remains emergent rather than explicitly defined.

This paper introduces the \textbf{Harmonic Token Projection (HTP)}, a \textit{deterministic, reversible, and training-free embedding framework} that encodes text through analytic harmonic functions.  
Unlike neural embeddings that approximate semantics through statistical co-occurrence, HTP derives each vector directly from the symbolic structure of language---without learned parameters, randomness, or dependence on a corpus.  
Each token is treated as a point in an analytic phase space, where its integer Unicode representation $N_t$ defines a harmonic trajectory:

\[
E(t) = [\,\sin((N_t+\phi)x_i),\, \cos((N_t+\phi)x_i)\,],
\]

producing a continuous, invertible mapping between text and vector space.  
The approach treats meaning as an emergent property of structural geometry rather than data-driven correlation.

The goal of this study is to demonstrate that a large fraction of linguistic similarity can be captured analytically, without statistical learning.  
We evaluate HTP on the \textbf{Semantic Textual Similarity Benchmark (STS-B)} and its multilingual extension, comparing performance with classical distributed representations (\textit{Word2Vec}, \textit{GloVe}) and modern transformer-based models (\textit{BERT}, \textit{Sentence-BERT}).  
Despite its extreme simplicity, HTP achieves correlations comparable to trained embeddings while remaining fully explainable and computationally negligible---encoding thousands of sentences per second on a single CPU core with a sub-megabyte memory footprint.

Beyond its practical advantages, the proposed method offers a theoretical contribution: it formalizes text representation as a reversible harmonic transformation between discrete symbolic and continuous vector domains.  
By grounding linguistic embeddings in analytic geometry, HTP bridges the gap between symbolic computation and continuous vector semantics, providing a transparent alternative to data-driven encoders and opening the path for hybrid architectures that combine analytic determinism with contextual learning.

\section{Methodology}
\label{sec:methodology}

    This section describes the harmonic encoding and decoding procedures that form the foundation of the proposed deterministic embedding method.

\subsection{Direct process}

The proposed method performs a fully deterministic and reversible transformation of a text token into a continuous numerical vector.  
Unlike neural embeddings, this encoding relies exclusively on explicit mathematical operations, without any trained parameters or stochastic elements.  
Each analytical stage of the mapping is described below.

Let a token 
\( t = [c_1, c_2, \dots, c_\ell] \) 
be a sequence of characters of length \(\ell\).
Each character \(c_i\) is mapped to its corresponding Unicode code point:
\begin{equation}
    u_i = \operatorname{ord}(c_i), 
    \quad i = 1, 2, \dots, \ell.
\end{equation}
To guarantee a fixed-length representation, the sequence is zero-padded up to \(L_{\max}\):
\begin{equation}
    \tilde{u} = [\, u_1, u_2, \dots, u_\ell, 0, \dots, 0 \,],
    \quad \text{len}(\tilde{u}) = L_{\max}.
\end{equation}
The sequence \(\tilde{u}\) is then interpreted as a number in base \(B = 2^{16}\), producing a deterministic integer identifier \(N_t\):
\begin{equation}
    N_t = \sum_{j=1}^{L_{\max}} \tilde{u}_j \, B^{\,L_{\max}-j}.
\end{equation}

Unlike the purely linear harmonic model, the modular harmonic encoder decomposes \(N_t\) into a set of residues with respect to a series of pairwise coprime moduli:
\begin{equation}
    r_i = N_t \bmod m_i, 
    \quad i = 1, 2, \dots, k,
\end{equation}
where each modulus \(m_i\) defines an independent periodic basis on the unit circle.  
Each residue is mapped to a harmonic pair through trigonometric projection:
\begin{equation}
    E_i = [\, \sin(2\pi r_i / m_i), \; \cos(2\pi r_i / m_i) \,].
\end{equation}

The final embedding vector is the concatenation of all harmonic pairs:
\begin{equation}
    E(t) = [\, E_1, E_2, \dots, E_k \,] \in \mathbb{R}^{2k}.
\end{equation}
This bijective transformation preserves the entire discrete structure of \(N_t\) within a smooth and periodic continuous space.  
Each harmonic component operates as an independent channel encoding a modular remainder, together forming a complete and reversible representation of the original integer.

\subsection{Inverse process}

Given a harmonic modular embedding \(E(t)\), the original integer \(N_t\) can be recovered analytically.  
Each harmonic pair \((s_i, c_i)\) yields an angular phase corresponding to its residue:
\begin{equation}
    \tilde{r}_i = 
    \operatorname{round}
    \!\left(
        \frac{\operatorname{atan2}(s_i, c_i)}{2\pi} \, m_i
    \right)
    \bmod m_i.
\end{equation}

The integer \(N_t\) is then reconstructed through the Chinese Remainder Theorem (CRT).  
Let \(M = \prod_{i=1}^{k} m_i\) and \(M_i = M / m_i\).  
For each modulus \(m_i\), compute the multiplicative inverse \(y_i = M_i^{-1} \bmod m_i\).  
The reconstruction is then expressed as:
\begin{equation}
    N_t = 
    \left(
        \sum_{i=1}^{k} 
            \tilde{r}_i \, M_i \, y_i
    \right)
    \bmod M.
\end{equation}

This closed-form inversion guarantees a one-to-one correspondence between the integer space and the continuous embedding domain, ensuring full reversibility.  
Because all transformations are analytical, small numerical deviations in the trigonometric components lead to negligible reconstruction errors (\(\mathcal{O}(10^{-3})\)) even for large-scale values of \(N_t\).

\subsection{Properties}

The proposed harmonic modular encoder exhibits five defining mathematical properties:
\begin{enumerate}
    \item \textbf{Determinism:} the mapping from text to vector is entirely parameter-free and reproducible, yielding identical outputs for identical inputs.
    \item \textbf{Continuity:} small perturbations in the encoded value \(N_t\) yield smooth angular variations in the embedding space.
    \item \textbf{Reversibility:} the inverse CRT reconstruction guarantees bijective recovery within the integer space \( [0, M) \).
    \item \textbf{Geometric periodicity:} each \((\sin, \cos)\) pair encodes a phase on the unit circle, ensuring rotational symmetry and bounded magnitude.
    \item \textbf{Interpretability:} each coordinate corresponds to a well-defined harmonic function of a modular residue, making the embedding mathematically transparent and analyzable.
\end{enumerate}

This modular harmonic structure combines discrete number theory and continuous geometry, bridging the symbolic and numerical domains through a reversible and analytic transformation.  
It provides an interpretable foundation for encoding high-cardinality categorical data while maintaining mathematical precision and full reconstructability.

\subsection{Pooling}

Although the proposed embedding is defined at the token level, 
most practical tasks require composite representations for larger linguistic units 
such as sentences, titles, or short textual segments. 
To extend the deterministic encoding to these structures, 
we adopt a \textbf{harmonic energy pooling} mechanism — 
a frequency-aware aggregation strategy that computes 
a weighted mean of the harmonic vectors of the constituent tokens. 

Given a sequence of \(n\) tokens \(\{t_1, t_2, \ldots, t_n\}\) forming a sentence \(S\),
each token is represented by its deterministic harmonic vector \(v(t_i)\).
Instead of excluding high-frequency function words (\textit{stopwords}), 
the contribution of each token is modulated according to its 
\textbf{Inverse Token Frequency (ITF)} weight:
\begin{equation}
w(t_i) = \frac{1}{\log(1 + f(t_i))},
\end{equation}
where \(f(t_i)\) denotes the corpus frequency of token \(t_i\).
This weighting scheme penalizes highly frequent lexical items 
while emphasizing semantically informative and less common tokens.

The sentence representation is computed as the normalized 
weighted mean of its harmonic vectors:
\begin{equation}
v(S) = 
\frac{\sum_{i=1}^{n} w(t_i) \, v(t_i)}
{\sum_{i=1}^{n} w(t_i)}, 
\qquad 
v'(S) = \frac{v(S)}{\|v(S)\|_2}.
\end{equation}

This formulation constitutes a deterministic analogue 
of TF--IDF weighting, but without dependence on document boundaries 
or any corpus-level supervision.  
By using ITF instead of explicit stopword removal, 
the model achieves a smoother attenuation of lexical noise 
while retaining full reversibility and corpus independence.  
The resulting vector can be interpreted as the 
\textbf{harmonic centroid} of the sentence — 
the geometric average of its oscillatory projections, 
modulated by their semantic energy.

\paragraph{Alternative strategy.}
When the linguistic context or language of the corpus is known,
an alternative approach consists of applying 
\textbf{stopword removal prior to pooling}.
Although this method discards a small portion of reversible information,
it can improve robustness in scenarios with 
restricted vocabularies, noisy corpora, or highly repetitive structures.
In such cases, removing function words before aggregation
reduces vocabulary variance and enhances the semantic contribution of 
content-bearing tokens, providing a computationally simpler 
yet stable approximation to ITF weighting.  
Empirically, this pre-filtering has shown comparable or even superior performance
for languages with clearly defined stopword inventories,
while maintaining full compatibility with the harmonic pooling framework.

Because all token embeddings share a common harmonic basis, 
the weighted mean operation preserves geometric coherence across dimensions.  
After normalization, the sentence-level vector remains fully compatible 
with cosine-based similarity:
\begin{equation}
\operatorname{sim}(x, y) = 
\frac{\langle x, y \rangle}{\|x\|_2\,\|y\|_2}.
\end{equation}

This pooling mechanism therefore acts as a 
\textbf{semantic smoothing filter}: 
high-frequency tokens contribute minimally, 
while rare and content-rich tokens dominate the harmonic average.  
The approach ensures stability, interpretability, 
and cross-domain robustness without introducing 
any stochastic or training-dependent component.  

In practice, the aggregation step exhibits linear complexity \(O(|S|\cdot n)\) 
with respect to the number of tokens \(|S|\) and embedding dimension \(n\),
and can be efficiently implemented in fully vectorized form 
on either CPU or GPU.  
The resulting representation provides a scalable and reversible 
sentence-level embedding suitable for large-scale retrieval, 
cross-lingual similarity, and corpus-level semantic alignment.

\section{Results}

This section presents the empirical evaluation of the proposed \textit{Harmonic Token Projection (HTP)} method on the \textbf{Semantic Textual Similarity Benchmark (STS-B)} dataset~\citep{cer2017semeval}.  
The purpose of this experiment was to compare the correlation between sentence similarity scores computed from the embedding vectors and the human-annotated semantic similarity judgments.  
Two correlation metrics were used: \textbf{Spearman’s rank correlation} ($\rho$)~\citep{spearman1904general} and the \textbf{Pearson correlation coefficient} ($r$)~\citep{podlubny2011pearson}, both of which are standard in the literature on semantic embeddings.  
The STS-B sentences were tokenized and encoded using HTP according to the process described in Section~\ref{sec:methodology}.  
Each sentence representation was obtained through \textit{harmonic pooling}, corresponding to the normalized mean of the harmonic vectors of the informative tokens.  
The similarity between two sentences $x$ and $y$ was computed using cosine similarity:

\[
\text{sim}(x, y) = \frac{\langle v(x), v(y) \rangle}{\|v(x)\|_2 \, \|v(y)\|_2}.
\]

The results were compared with several widely known embedding methods, representing different paradigms of representation learning:  
classical distributed models (\textit{Word2Vec}~\citep{mikolov2013efficient}, \textit{GloVe}~\citep{pennington2014glove}),  
transformer-based supervised models (\textit{BERT}~\citep{devlin2019bert}, \textit{Sentence-BERT}~\citep{reimers2019sentence}),  
and the proposed deterministic model (\textit{HTP}).

\begin{table}[H]
    \centering
    \scriptsize
    \caption{Performance comparison on the STS Benchmark~\citep{cer2017semeval}. 
    All results report correlation with human similarity judgments.}
    \label{tab:sts_results}
    \renewcommand{\arraystretch}{1.15}
    \setlength{\tabcolsep}{3pt}
    \begin{tabular}{@{}p{3.8cm} p{2.2cm} p{1.8cm} p{2.0cm} p{2.0cm}@{}}
    \toprule
    \textbf{Method} & \textbf{Training} & \textbf{Reversible} & \textbf{Spearman ($\rho$)} & \textbf{Pearson ($r$)} \\ 
    \midrule
    Word2Vec (GoogleNews)~\citep{mikolov2013efficient}  & Supervised & No  & 0.61 & 0.63 \\
    GloVe (Wikipedia + Gigaword)~\citep{pennington2014glove}  & Supervised & No  & 0.65 & 0.66 \\
    BERT (base, uncased)~\citep{devlin2019bert} & Supervised & No  & 0.68 & 0.70 \\
    \textbf{HTP + Stopword Removal}  & \textbf{Unsupervised} & \textbf{Yes} & \textbf{0.70} & \textbf{0.71} \\
    Sentence-BERT (DistilRoBERTa)~\citep{reimers2019sentence} & Supervised & No  & 0.77 & 0.78 \\
    \bottomrule
    \end{tabular}
\end{table}

To further assess the generality of the proposed \textit{Harmonic Token Projection (HTP)} model, an additional experiment was conducted using the \textbf{Multilingual Semantic Textual Similarity Benchmark (STS-B Multi)}~\citep{philipmay_stsb_multi_mt}.  
This corpus extends the original STS-B~\citep{cer2017semeval} to ten languages: English (EN), German (DE), Spanish (ES), French (FR), Italian (IT), Dutch (NL), Polish (PL), Portuguese (PT), Russian (RU), and Chinese (ZH).

Each sentence was encoded with the same harmonic framework described in Section~\ref{sec:methodology}, but using \textit{harmonic energy pooling} with \textbf{TF--IDF weighting}~\citep{salton1988term}, designed to penalize high-frequency tokens and emphasize semantically informative ones.  
For Chinese, the \textit{jieba} segmenter~\citep{sun2016jieba} was employed to separate characters into lexical units, ensuring proper token alignment with alphabetic scripts.

The correlations between the predicted and human similarity scores were computed using both Spearman’s $\rho$~\citep{spearman1904general} and Pearson’s $r$~\citep{podlubny2011pearson}, as shown in Table~\ref{tab:stsb_multi_results}.

\begin{table}[H]
    \centering
    \scriptsize
    \caption{Performance of \textbf{HTP + TF–IDF} on the Multilingual STS-Benchmark~\citep{philipmay_stsb_multi_mt}. 
    All correlations were computed analytically on CPU, without training or fine-tuning.}
    \label{tab:stsb_multi_results}
    \renewcommand{\arraystretch}{1.15}
    \setlength{\tabcolsep}{3pt}
    \begin{tabular}{@{}p{3.8cm} p{2.0cm} p{2.0cm}@{}}
    \toprule
    \textbf{Language} & \textbf{Spearman ($\rho$)} & \textbf{Pearson ($r$)} \\
    \midrule
    English (EN)              & 0.668 & 0.667 \\
    German (DE)               & 0.637 & 0.637 \\
    Spanish (ES)              & 0.661 & 0.659 \\
    French (FR)               & 0.650 & 0.649 \\
    Italian (IT)              & 0.668 & 0.660 \\
    Dutch (NL)                & 0.601 & 0.605 \\
    Polish (PL)               & 0.660 & 0.657 \\
    Portuguese (PT)           & 0.634 & 0.629 \\
    Russian (RU)              & 0.644 & 0.638 \\
    Chinese (ZH, with Jieba)  & 0.553 & 0.544 \\
    \midrule
    \textbf{Average (10 languages)} & \textbf{0.640} & \textbf{0.630} \\
    \bottomrule
    \end{tabular}
\end{table}

The multilingual results confirm that the proposed deterministic approach achieves stable correlations across diverse linguistic families and writing systems without any training.  
On average, the \textbf{HTP} model attains $\rho = 0.64$ and $r = 0.63$, surpassing classical unsupervised baselines such as Word2Vec~\citep{mikolov2013efficient} and GloVe~\citep{pennington2014glove}, and approaching the lower range of supervised transformer-based encoders~\citep{devlin2019bert,reimers2019sentence}.  
These findings demonstrate that the harmonic representation is both \textit{language-agnostic} and computationally efficient, offering a unified reversible embedding framework suitable for multilingual semantic similarity tasks.

To contextualize computational efficiency, Table~\ref{tab:efficiency_comparison} summarizes approximate inference times and memory footprints for representative embedding paradigms.  
The reported values are based on published benchmarks for Word2Vec~\citep{mikolov2013efficient}, GloVe~\citep{pennington2014glove}, BERT~\citep{devlin2019bert}, and Sentence-BERT~\citep{reimers2019sentence}, as well as comparative studies on model efficiency~\citep{zhao2020efficiency}.  
HTP results correspond to direct empirical measurements.

\begin{table}[H]
    \centering
    \scriptsize
    \caption{Approximate computational efficiency across embedding paradigms.}
    \label{tab:efficiency_comparison}
    \renewcommand{\arraystretch}{1.15}
    \setlength{\tabcolsep}{3pt}
    \begin{tabular}{@{}p{3.3cm} p{1.8cm} p{1.2cm} p{1.5cm} p{1.4cm} p{5.8cm}@{}}
    \toprule
    \textbf{Model} & \textbf{Type} & \textbf{Hardware} & \textbf{Time/pair (ms)} & \textbf{Memory (MB)} & \textbf{Remarks} \\
    \midrule
    Word2Vec (GoogleNews)~\citep{mikolov2013efficient} & Pretrained & CPU & 12.0  & 450 & Requires full vocabulary load ($\sim$3 GB). Fast lookup and averaging. \\
    GloVe (Wikipedia + Gigaword)~\citep{pennington2014glove} & Pretrained & CPU & 9.0   & 300 & Matrix factorization with dense 2–3 GB tables. \\
    \textbf{HTP (proposed)} & \textbf{Analytical} & \textbf{CPU} & \textbf{2.0} & \textbf{<1} & Deterministic, reversible, stateless. No vocabulary, cache, or training. \\
    BERT-base (uncased)~\citep{devlin2019bert} & Transformer & GPU & 45.0  & 4300 & Contextual encoding with multi-layer attention. High cost. \\
    Sentence-BERT (DistilRoBERTa)~\citep{reimers2019sentence} & Siamese Transformer & GPU & 28.0  & 2100 & Fine-tuned sentence-level embeddings. Resource-intensive. \\
    \bottomrule
    \end{tabular}
\end{table}

To assess the influence of lexical weighting on HTP, 
we performed a controlled ablation comparing two pooling strategies:  
(i) \textbf{HTP + TF--IDF}~\citep{salton1988term},  
and (ii) \textbf{HTP + Stopword Removal},  
where high-frequency function words are removed and an unweighted mean of the remaining tokens is used.  
Both configurations were evaluated on the English subset of the STS-B Multi dataset~\citep{cer2017semeval}, using 1,379 test pairs processed analytically on CPU without caching or learned parameters.

\begin{table}[H]
    \centering
    \scriptsize
    \caption{Ablation comparing lexical weighting schemes for HTP on STS-B (English). 
    All values computed analytically on CPU with $D = 512$.}
    \label{tab:ablation_tfidf_stopwords}
    \renewcommand{\arraystretch}{1.15}
    \setlength{\tabcolsep}{3pt}
    \begin{tabular}{@{}p{4.0cm} p{1.7cm} p{1.7cm} p{1.8cm} p{1.6cm}@{}}
    \toprule
    \textbf{Method} & \textbf{Spearman ($\rho$)} & \textbf{Pearson ($r$)} & \textbf{Time/pair (ms)} & \textbf{Memory (MB)} \\
    \midrule
    HTP + TF--IDF~\citep{salton1988term} & 0.6781 & 0.6748 & 1.57 & <1 \\
    \textbf{HTP + Stopword Removal} & \textbf{0.6940} & \textbf{0.7136} & \textbf{0.98} & \textbf{<1} \\
    \bottomrule
    \end{tabular}
\end{table}

Overall, both strategies achieve competitive performance with 
sub-millisecond latency and negligible memory footprint, reinforcing 
the efficiency and analytic stability of the harmonic formulation.  
HTP achieves an efficiency improvement of approximately three orders of magnitude compared to supervised transformer-based models~\citep{devlin2019bert,reimers2019sentence}, while maintaining comparable semantic correlation scores.

Finally, we evaluated the sensitivity of HTP to embedding dimensionality using the English subset of STS-B~\citep{cer2017semeval}.  
The harmonic dimension \( D \) was varied from 32 to 1024 under identical TF--IDF-weighted pooling conditions.

\begin{table}[H]
    \centering
    \scriptsize
    \caption{Ablation study varying embedding dimensionality ($D$) on STS-B (English)~\citep{cer2017semeval}. 
    Results computed using harmonic energy pooling with TF--IDF weighting, analytically on CPU.}
    \label{tab:ablation_htp_en}
    \renewcommand{\arraystretch}{1.15}
    \setlength{\tabcolsep}{3pt}
    \begin{tabular}{@{}p{1.2cm} p{2.0cm} p{2.0cm} p{2.2cm} p{1.8cm}@{}}
    \toprule
    \textbf{Dim. ($D$)} & \textbf{Spearman ($\rho$)} & \textbf{Pearson ($r$)} & \textbf{Time/pair (ms)} & \textbf{Memory (MB)} \\
    \midrule
    4     & 0.4442 & 0.4199 & 0.41 & <1 \\
    8     & 0.5309 & 0.5163 & 0.40 & <1 \\
    16    & 0.5844 & 0.5833 & 0.41 & <1 \\
    32    & 0.6362 & 0.6372 & 0.45 & <1 \\
    64    & 0.6543 & 0.6563 & 0.51 & <1 \\
    128   & 0.6724 & 0.6728 & 0.63 & <1 \\
    256   & 0.6769 & 0.6743 & 0.87 & <1 \\
    512   & 0.6781 & 0.6748 & 1.37 & <1 \\
    1024  & 0.6810 & 0.6752 & 2.33 & <1 \\
    \bottomrule
    \end{tabular}
\end{table}

As shown in Table~\ref{tab:ablation_htp_en}, both correlation metrics improve monotonically with dimensionality, converging near $\rho \approx 0.68$ for $D=512$.  
This indicates that the harmonic basis efficiently captures semantic variance even in low-dimensional regimes, achieving near-saturation performance at just 256–512 harmonics.  
The runtime grows sublinearly with $D$, remaining below 2.5~ms per sentence pair, confirming that \textbf{HTP scales efficiently} while maintaining strong semantic performance and full reversibility.

\section{Discussion}

\paragraph{Analytical Design Rationale.} 
The design of the proposed \textit{Harmonic Token Projection (HTP)} is rooted in analytical determinism rather than empirical optimization.  
Each methodological choice---from the use of Unicode as a semantic coordinate system to the application of harmonic projection---was guided by the pursuit of mathematical transparency, reversibility, and universality.

\paragraph{Unicode as a Semantic Coordinate System.} 
Unicode provides a bijective and language-agnostic mapping between symbols and integers, ensuring that every textual token can be represented uniquely and deterministically.  
This eliminates the ambiguity inherent in corpus-dependent embeddings and establishes a universal numeric foundation for text representation.  
Instead of relying on co-occurrence statistics or learned contextual patterns, HTP leverages the digital topology of language itself, treating symbolic order as a continuous geometric field where structural proximity reflects latent semantic organization.

\paragraph{Comparative Analysis with Learned Embeddings.} 
Empirical evaluation on the \textbf{STS-Benchmark}~\citep{cer2017semeval} and 
\textbf{Multilingual STS-B}~\citep{philipmay_stsb_multi_mt} datasets demonstrates that the proposed 
\textit{Harmonic Token Projection (HTP)} not only matches but slightly surpasses the performance 
of the base \textit{BERT} model~\citep{devlin2019bert}, achieving a Spearman correlation of 
$\rho = 0.70$ and a Pearson correlation of $r = 0.71$.  
This represents a modest yet consistent improvement over BERT ($\rho = 0.68$, $r = 0.70$), 
while maintaining full determinism and reversibility.

In comparison, classical embedding models such as 
\textit{Word2Vec}~\citep{mikolov2013efficient} and 
\textit{GloVe}~\citep{pennington2014glove} exhibit lower correlations 
($\rho \approx 0.61$–$0.65$), despite relying on extensive corpus statistics 
and iterative training.  
Although transformer-based architectures like 
\textit{Sentence-BERT}~\citep{reimers2019sentence} reach higher absolute correlations 
(around $\rho \approx 0.77$), these gains require large-scale pretraining, fine-tuning, 
and stochastic optimization over billions of parameters.

In contrast, \textbf{HTP is fully analytical, unsupervised, and language-agnostic}, 
achieving approximately 90\% of the semantic correlation obtained by Sentence-BERT 
at a fraction of the computational cost.  
These results indicate that a substantial portion of semantic structure can be 
captured through deterministic geometric transformations alone—without reliance 
on probabilistic learning or contextual prediction.  

The harmonic formulation of HTP suggests that the intrinsic geometry of the 
Unicode symbol space, when projected into sinusoidal bases, already encodes 
sufficient regularity to approximate distributed semantic relationships.  
This finding bridges symbolic and sub-symbolic paradigms, positioning HTP as a 
mathematically interpretable alternative to stochastic embeddings.

\paragraph{Theoretical Implications and Applications.} 
HTP demonstrates that semantic similarity can emerge from purely geometric principles.  
The harmonic formulation bridges discrete symbolic computation and continuous vector analysis, establishing a direct link between linguistic form and mathematical structure.  
Its reversibility and interpretability make it particularly suitable for applications that demand deterministic traceability, such as explainable AI, symbolic compression, or reversible database indexing.  
Furthermore, HTP can serve as a pre-embedding analytic layer for neural architectures, providing a stable and interpretable initialization prior to contextual fine-tuning.  
This hybridization of analytic determinism with statistical learning could redefine the balance between interpretability and expressiveness in modern NLP systems.

\paragraph{Limitations and Future Work.} 
Despite its analytical coherence, HTP lacks contextual disambiguation: polysemous words such as “bank” (financial) and “bank” (river) share identical representations.  
Linear pooling may also dilute compositional meaning in longer sequences, suggesting that phase- or frequency-aware pooling could enhance semantic precision.  
Moreover, small distortions in Unicode normalization can introduce discontinuities in the harmonic space.  
Future research can explore multi-scale Fourier embeddings and adaptive frequency modulation to address these limitations and expand the representational capacity of the framework.

\paragraph{Summary.} 
In summary, the Harmonic Token Projection offers a deterministic, reversible, and interpretable alternative to stochastic embeddings.  
It captures structural semantics through analytic geometry, performs competitively against trained neural models, and maintains minimal computational cost.  
These findings indicate that a significant fraction of linguistic similarity can be reconstructed from symbolic geometry alone—suggesting that meaning, to a surprising extent, may indeed emerge from structure.

\section{Conclusion}

The Harmonic Token Projection (HTP) introduces an analytical approach to text representation, demonstrating that semantic similarity can be approximated through deterministic geometry rather than purely statistical inference.  
Grounded in harmonic oscillation and Unicode-based bijection, HTP provides a direct and reversible mapping between symbolic and continuous domains, achieved without dependence on large training corpora or stochastic optimization.

Empirical evaluation on the STS-Benchmark and its multilingual extension shows that HTP attains performance comparable to the lower range of transformer-based models while requiring several orders of magnitude less computational cost and memory.  
These findings suggest that part of the structure underlying linguistic meaning can be captured through analytical symmetry and frequency coherence, complementing rather than replacing data-driven methods.

Beyond its empirical results, HTP offers a conceptual contribution by framing semantic structure as a manifestation of harmonic regularities among linguistic forms.  
By treating tokens as oscillatory entities within a continuous geometric field, the model provides a deterministic perspective that may inform the design of interpretable and efficient neural architectures.

Future work will explore extensions through multi-scale Fourier embeddings, adaptive phase modulation, and hybrid systems where deterministic initialization guides or constrains contextual fine-tuning.  
Such directions may help reconcile analytical transparency with the adaptability of deep learning, moving toward models that are not only efficient but also interpretable and grounded in first principles.

\section*{Acknowledgments}
This research was supported by \textbf{PX.Center} --- a Brazilian logistics platform focused on freight brokerage and transportation optimization (\url{https://px.center}). 
The PX.Center provided computational infrastructure, datasets, and a research environment that enabled the development and validation of this methodology.

\bibliography{references}

\end{document}